# Det-SLAM: A semantic visual SLAM for highly dynamic scenes using Detectron2

Ali Eslamian
*Department of Electrical and Computer Engineering*
Isfahan University of Technology
Isfahan, Iran
a.eslamian@ec.iut.ac.ir

Mohammad Reza Ahmadzadeh
*Department of Electrical and Computer Engineering*
Isfahan University of Technology
Isfahan, Iran
Ahmadzadeh@iut.ac.ir

*Abstract*— According to experts, Simultaneous Localization and Mapping (SLAM) is an intrinsic part of autonomous robotic systems. Several SLAM systems with impressive performance have been invented and used during the last several decades. However, there are still unresolved issues, such as how to deal with moving objects in dynamic situations. Classic SLAM systems depend on the assumption of a static environment, which becomes unworkable in highly dynamic situations. Several methods have been presented to tackle this issue in recent years, but each has its limitations. This research combines the visual SLAM systems ORB-SLAM3 and Detectron2 to present the Det-SLAM system, which employs depth information and semantic segmentation to identify and eradicate dynamic spots to accomplish semantic SLAM for dynamic situations. Evaluation of public TUM datasets indicates that Det-SLAM is more resilient than previous dynamic SLAM systems and can lower the estimated error of camera posture in dynamic indoor scenarios.

## I. Introduction

In the majority of SLAM algorithms, indoor or confined areas are prioritized. It is usually considered that these locations include no moving objects. Although this assumption is impractical in the real world, several approaches exist to solve this problem. The first solution uses geometrical methods to disregard moving components and create a structural map based on the static area. The second solution uses deep neural networks to detect specific moving objects and exclude them from the post-processing algorithm.

Moreover, semantic information is essential for the robot to comprehend its workplace environment as deep learning advances; specific networks may achieve more remarkable semantic segmentation. Combining these networks with SLAM could create a semantic map, enhancing the robot's perception. The third option uses image processing techniques such as morphology or background subtractor to remove and replace moving elements with a static scene. These strategies have advantages and disadvantages and can be merged to build a new algorithm compatible with various circumstances.

## II. Related Works

Visual SLAM algorithms generally accept monocular, stereo, or RGB-D camera input. In recent researches, an event-based camera has been deployed to improve motion blur performance. These algorithms use a variety of approaches to extract data from images to generate maps for the navigator.

### A. SLAM Architectures

Assuming that the environment is stationary, existing algorithms can be categorized into two subgroups. In the indirect method, the illumination of each pixel in each frame is compared to the next frame; then, correlation is computed to generate the map. Based on correlation correspondence, it is categorized into dense, semi-dense, and sparse maps. This approach is more resilient to motion blur, although it is computationally expensive and susceptible to pixel noise. DTAM [1] is one of these algorithms that produce dense maps. LSD-SLAM [2] is a real-time technique that generates a semi-dense environment map but does not support loop closure. DSOD [3] is another example of an algorithm that produces sparse maps, especially in dynamic environments.

In the feature-based method, the algorithm extracts and analyses only specific feature points from each frame. It is preferred over the direct technique due to its lower computational costs. Mono-SLAM [4] is a feature-based algorithm that uses a filter to track landmarks and estimate position. However, a significant number of landmarks cause higher computational costs. ORB-SLAM2 [5], one of the most successful feature-based algorithms, consists of three main threads for tracking and mapping and is the basis for many subsequent feature-based algorithms. These algorithms are susceptible to many failures in dynamic situations but despite modifications, they may be used in later algorithms.

### B. Semantic SLAM

DS-SLAM [6] is a further robust algorithm based on ORB-SLAM2 with two additional threads. This algorithm reduces the impact of dynamic objects in vision-based SLAM by combining the semantic segmentation network with the optical flow approach while presenting the octo-tree map semantically. In SegNet [7], known dynamic objects are first recognized; subsequently, a moving consistency check is performed on selected feature points of the segmented region. If it is determined that the mass of key points is dynamic the entire object's feature points are ignored in the next steps.

DynaSLAM [8] is a system for visual SLAM with dynamic object recognition and inpainting. Using Mask-RCNN [9], this system detects and eliminates a priori dynamic items. Additionally, multi-view geometry assists in locating unidentified dynamic objects. The algorithm uses the depth image of a previously segmented object to inpaint the desired area. Then it estimates a map of the static components of the scene, which is required for real-world applications. Next, the key points are extracted using the ORB-SLAM2 feature extraction technique. Finally, it computes the angle of parallax for each key point. Therefore, immobile objects with parallax angles exceeding 30 degrees were deemed dynamic due to their viewpoint difference.

The general performance of DP-SLAM [10] has been developed from Dyna-SLAM. For semantic segmentation, this algorithm employs Mask R-CNN. Additionally, epipolar geometric approaches like DS-SLAM exclude misclassified feature points. Since the highest error probability occurs at the segmentation's boundary edges, the algorithm determines the probability of moving points based on the previous frames and Bayes' law. Therefore, the feature point is discarded from further processing if it is likely to exceed the threshold.

YOLO-SLAM [11] is based on ORB-SLAM2 but consists of two extra threads, semantic segmentation and dynamic feature screening. First, the segmentation thread uses a modified version of Darknet19-YOLOv3 to select the most known dynamic objects. The dynamic feature screening thread uses geometry depth RANSAC to divide feature points into dynamic and static subgroups. This method calculates the depth variance of feature points in each bounding box to distinct statice points from dynamic ones.

PSPNet-SLAM [12] is an example of a practical algorithm. Similar to DS-SLAM, this approach employs semantic segmentation and optical flow to recognize and classify moving entities. This method differs from DS-SLAM in that it uses PSPNet instead of SegNet for segmentation. The improved RANSAC model is also used to eliminate outliers.

Another recently proposed algorithm is Blitz-SLAM [13] which is also based on ORB-SLAM2. There are two more threads, semantic segmentation using BlitzNet and a geometrical component. Because of this network's segmentation inaccuracies, depth information is employed to change the segmentation mask. After proper segmentation and separation of moving elements from the background, the epipolar condition of the stationary background is evaluated, and the point cloud is prepared.

III. SYSTEM OVERVIEW

Our approach is based on ORB-SLAM3 [14], which has outstanding performance in practical environments. Also, it is the last version of the ORB feature-based concurrent SLAM system. Building a semi-dense map from inputs that follow three main parallel threads. Although it fails in a highly dynamic situation, it performs effectively in a static one. We have included a semantic segmentation thread in our technique to detect and remove each dynamic item in frames. Consequently, moving components are eliminated prior to further processing. In addition, we modified a part of the depth image input using image processing techniques, which helps to recognize previously undetected moving objects. It allows the algorithm to operate more quickly than comparable algorithms using geometrical approaches. We used Detectron2 [15] in preprocessing of ORB-SLAM3; So it is called Det-SLAM. Our algorithm consists of two main parts:

A. Semantic Segmentation

In this work, we apply the state-of-the-art object detection with Detectron2 in order to detect dynamic objects. Detectron2 is the most recent library for generation represented by the Facebook AI Research group (FAIR). It contributes effectively to several computer vision research projects and Facebook applications. The backbone network, Region Proposal Network (RPN), and box head are the three fundamental architectural components of Detectron2. The backbone network design utilizes the Feature Pyramid Network (FPN) to extract feature maps from input images of varying scales. In the RPN, features are extracted on various scales with a confidence score, and the Region of Interest (ROI) of an item is cropped in box head.

We employ pre-train weights of COCO objects detection baselines for an instant segmentation module to identify dynamic objects, such as humans, who are primarily mobile. ResNet and FPN backbones are used with convolutional layers and FCN heads to predict masks and boxes, respectively. These items' boundaries will be determined, and the region will be highlighted for further processing. Detectron2 utilizes the bounding box to define the spatial location of an object and independently sketches the object's boundary.

However, segmentation generates image edges, which in turn creates ORB features; the method removes any ORB features that fall inside the border of an item. Correspondingly, the key points located in that region will be omitted. Other ORB features included inside the bounding box would be candidates for the next stage.

B. Depth Processing

A depth map corresponding to a specific ROI in the RGB image determined by bounding boxes is prepared in the following step. At this stage, the depth information of all areas within the image's bounding box is considered. ORB features placed inside the object's depth range in the ROI are excluded from further processing.

All ROI with approximately the same depth of moving objects are eliminated to remove undetected suspicious objects affected by a moving factor, even though some data of static points will be lost.

Let $R^t$ represents the depth range of information in ROI at time $t$ and $P_j^t$ represents a pixel with the depth of $R_i^t$. The minimum and maximum value of depth information in the object area would be defined as

$$m_{obj} = \min_{object\ area}\{R_i^t\} \quad (1)$$

$$M_{obj} = \max_{object\ area}\{R_i^t\} \quad (2)$$

and maximum and minimum depth values of ROI define as

$$m_{ROI} = \min_{ROI}\{R_i^t\} \quad (3)$$

$$M_{ROI} = \max_{ROI}\{R_i^t\} \quad (4)$$

So, we define the depth difference of ROI as

$$d = M_{ROI} - m_{ROI} \quad (5)$$

We define a coefficient $\alpha$ to determine ratio of depth difference. So, all pixels of $P_j^t$ in ROI with the depth range in

$$m_{obj} - \alpha.d \leqslant R_i^t \leqslant M_{obj} + \alpha.d \quad (6)$$

will be removed from further processing. Determining the value of $\alpha$ depends on the image texture. The higher contrast in image illumination, the lower the $\alpha$ value. Since objects are continuous, the depth information of objects is in a confined range. It would be possible that the object's boundaries might exceed the object in the instance segmentation, so the depth information of some regions might have some abrupt change in the boundaries. Therefore, we only consider pixels with the maximum number of the same depth information. Fig. 1 shows overall view system components for Det-SLAM.

## EVALUATIONS AND RESULTS

In this section, experimental outcomes demonstrating the effectiveness of Det-SLAM are shown. Using the public TUM RGB-D dataset [16], we compared our technique against methods within the same category. Even though our technique is not real-time and the semantic segmentation thread must be prepared prior to feature extraction, the optimal speed/accuracy balance is accomplished. The TUM RGB-D dataset contains a number of sequences containing low and high dynamic range scenarios, which are assessed under comparable conditions. We employ Absolute Trajectory Error (ATE), which compares the projected trajectory to the ground truth at each time stamp. Due to removing some part of input images, RANSAC algorithm does not have ability to match the newly extracted ORB features with the previous one since corresponding ORB features in current frame might be lost in the next frame. So, trajectory has been lost due to the loss of key points in some frames. To further validate the evaluation, the trajectory proportion of matched pairs is added. Table 1 shows the mean, median, and standard deviation for the remaining ground truth and output pairings. In addition, Table 1 illustrates the ATE value of visual dynamic SLAM algorithms, such as DS-SLAM, Dyna-SLAM, DP-SLAM, Blitz-SLAM, PSPNet-SLAM, and YOLO-SLAM, for evaluating the accuracy of detail trajectory in dynamic situations. There are five most known sequences with dynamic object evaluated in mentioned SLAM algorithms. The "f/s/static" sequence shows two people gesticulating for a while. However, they are not moving around the room, our algorithm detect them as dynamics. The "f/w/static" is the same scene with two walking people. Also, the "f/w/half" is the same scene but camera moved around a half of sphere. The "f/w/xyz" and "f/w/rpy" sequences are from the same scene too but the camera moved along around principle axis for roll-pitch-yaw and along xyz axis respectively. The last four sequences represent the highly dynamic scenes. Table 1 confirms that the results for the "f/w/static" and "f/s/static" sequences are optimized in comparison of other methods. The common aspect of these two sequences is the stability of camera as we did not modify tracking part of the Det-SLAM algorithm.

The Relative Pose Error (RPE), which determines the relationship between two frame sequences, is depicted in Table 2.

Fig. 2 depict the ATE and RPE for each dataset. It can be seen trajectory drift time is not significant in all sequences. All studies are conducted on a computer with an Intel Core *i*7, GeForce GTX 1650 GPU, and 16GB of RAM.

## CONCLUSION

In this study, we proposed a new algorithm, Det-SLAM, for object detection and semantic segmentation by developing Detectron2. Detectron2 is more accurate than other comparable modules. Relations between class labels in semantic segmentation with fully connected networks, might be lost. Additionally, it may ignore small items or recognize huge objects sporadically. Det-SLAM makes the existing SLAM algorithms more robust and lessens the impact of dynamic objects on posture estimation. Although our approach achieved successful performance with TUM RGB-D datasets, it had some defects in some sequences compared to other comparable algorithms. We substituted image processing techniques instead of geometrical constraints in comparable algorithms. So the number of computations is minimized. Although we did not compare the execution time of the algorithm to that of other methods, our approach can be executed on standard PC desktops. From another viewpoint, in this procedure, we discarded a great deal of visual information based on the conditions, which led to tracking frame mistakes and missing some sequences. So, it can be claimed that the system is inapplicable to outdoor environments or rapid camera movements.

This algorithm provides several benefits over comparable methods mentioned before. The SegNet used in DS-SLAM is not susceptible to object overlap. It leads to errors in the final result. However, this does not pose a significant issue for the proposed algorithm. Furthermore, we used ORB-SLAM3 instead of ORB-SLAM2 which optimizes key frame processing more effectively. It also has the potential to use the IMU section in the future. Future developments of this work might include real-time performance, IMU data to compensate for missing information, and more image processing approaches for separating the foreground from the input image.

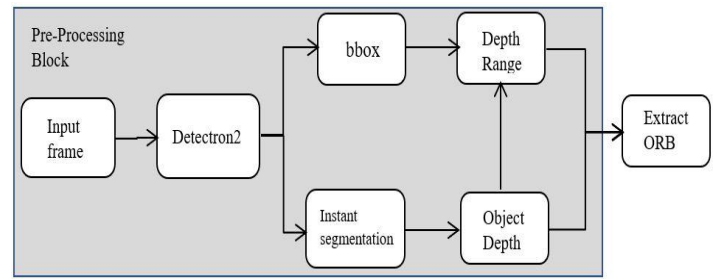

Fig.1 Pre-Processing frameworks of Det-SLAM

TABEL 1. Results of Absolute Trajectory Error Metrics

| sequences | | f/w/xyz | f/w/static | f/w/rpy | f/w/half | f/s/static |
|---|---|---|---|---|---|---|
| DS-SLAM | RMSE | 0.0247 | 0.0081 | 0.4442 | 0.0303 | 0.0065 |
| | Mean | 0.0186 | 0.0073 | 0.3768 | 0.0258 | 0.0055 |
| | Median | 0.0151 | 0.0067 | 0.2835 | 0.0222 | 0.0049 |
| | S.D. | 0.0161 | 0.0036 | 0.2350 | 0.0159 | 0.0033 |
| Dyna-SLAM | RMSE | 0.0156 | 0.0068 | 0.0417 | 0.0301 | 0.0063 |
| | Mean | 0.0134 | 0.0059 | 0.0312 | 0.0258 | 0.0055 |
| | Median | 0.0118 | 0.0052 | 0.0240 | 0.0218 | 0.0049 |
| | S.D. | 0.0079 | 0.0034 | 0.0275 | 0.0155 | 0.0031 |
| DP-SLAM | RMSE | **0.0141** | 0.0079 | 0.0356 | **0.0254** | 0.0059 |
| | Mean | **0.0120** | 0.0070 | 0.0277 | **0.0219** | 0.0051 |
| | Median | **0.0106** | 0.0063 | 0.0224 | **0.0183** | 0.0047 |
| | S.D. | 0.0073 | 0.0037 | 0.0218 | 0.0129 | 0.0029 |
| Blitz-SLAM | RMSE | 0.0153 | 0.0102 | 0.0356 | 0.0256 | - |
| | Mean | - | - | - | - | - |
| | Median | - | - | - | - | - |
| | S.D. | 0.0078 | 0.0052 | 0.0220 | **0.0126** | - |
| PSPNet-SLAM | RMSE | 0.0156 | 0.0072 | **0.0333** | 0.0255 | 0.0058 |
| | Mean | 0.0135 | 0.0064 | **0.0261** | 0.0222 | 0.0050 |
| | Median | 0.0117 | 0.0058 | **0.0204** | 0.0196 | 0.0044 |
| | S.D. | 0.0078 | 0.0033 | **0.0206** | 0.0126 | 0.0029 |
| YOLO-SLAM | RMSE | 0.0146 | 0.0073 | 0.2164 | 0.0283 | 0.0066 |
| | Mean | - | - | - | - | - |
| | Median | - | - | - | - | - |
| | S.D. | **0.0070** | 0.0035 | 0.1001 | 0.0138 | 0.0033 |
| Det-SLAM (Ours) | RMSE | 0.0482 | **0.0017** | 0.0389 | 0.0925 | **0.0036** |
| | Mean | 0.0383 | **0.0012** | 0.0309 | 0.0739 | **0.0031** |
| | Median | 0.0329 | **0.0008** | 0.0256 | 0.0631 | **0.0030** |
| | S.D. | 0.0296 | **0.0012** | 0.0237 | 0.0557 | **0.0017** |
| | Traj | 35.12% | 50.06% | 48.63% | 47.99% | 49.66% |

TABEL 2. Results of Relative Pose Error Metrics of our Det-SLAM

| Sequence | f/w/xyz | f/w/static | f/w/rpy | f/w/half | f/s/static |
|---|---|---|---|---|---|
| RMSE | 0.0653 | 0.0100 | 0.0680 | 0.01674 | 0.0223 |
| Mean | 0.0517 | 0.0074 | 0.0551 | 0.1310 | 0.0180 |
| Median | 0.0450 | 0.0065 | 0.0478 | 0.1045 | 0.0141 |
| S.D. | 0.0398 | 0.0067 | 0.0398 | 0.1042 | 0.0132 |

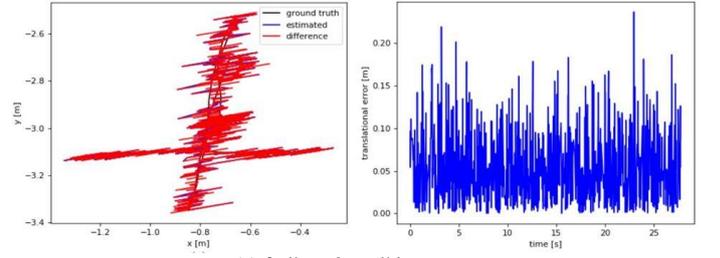

(a) freiburg3_walking_xyz

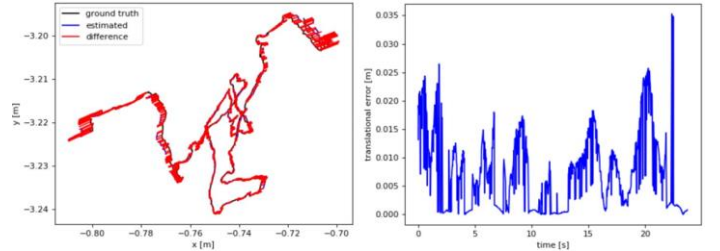

(b) freiburg3_walking_static

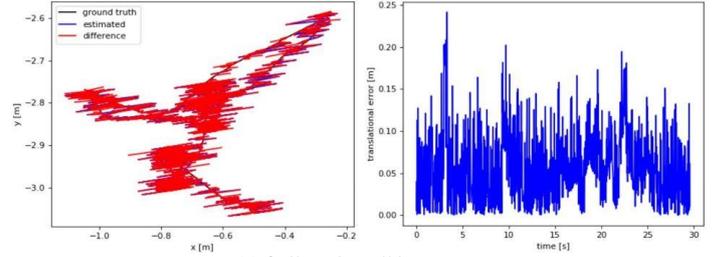

(c) freiburg3_walking_rpy

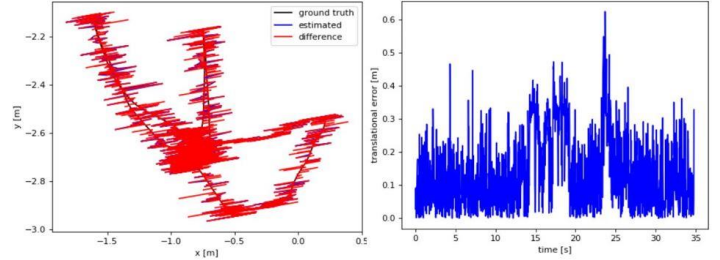

(d) freiburg3_walking_halfsphere

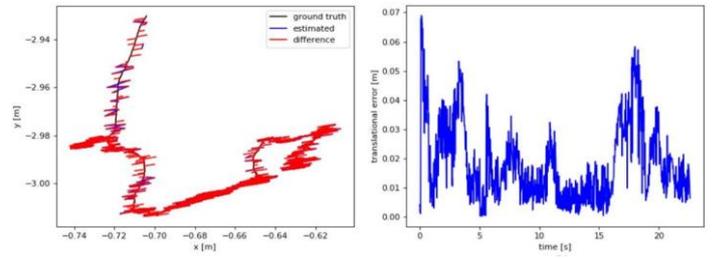

(e) freiburg3_sitting_static

Fig. 2 Result of five TUM RGB-D sequences in our Det-SLAM algorithm.
(Left) ATE based on estimated and ground truth two dimensional trajectory.
(Right) RPE with scale of time and error distance

## REFERENCES

[1] R. A. Newcombe, S. J. Lovegrove, and A. J. Davison, "DTAM: Dense tracking and mapping in real-time," in *2011 international conference on computer vision*, 2011: IEEE, pp. 2320-2327.

[2] J. Engel, T. Schöps, and D. Cremers, "LSD-SLAM: Large-scale direct monocular SLAM," in *European conference on computer vision*, 2014: Springer, pp. 834-849.

[3] P. Ma, Y. Bai, J. Zhu, C. Wang, and C. Peng, "DSOD: DSO in dynamic environments," *IEEE Access,* vol. 7, pp. 178300-178309, 2019.

[4] A. J. Davison, I. D. Reid, N. D. Molton, and O. Stasse, "MonoSLAM: Real-time single camera SLAM," *IEEE transactions on pattern analysis and machine intelligence,* vol. 29, no. 6, pp. 1052-1067, 2007.

[5] R. Mur-Artal and J. D. Tardós, "Orb-slam2: An open-source slam system for monocular, stereo, and rgb-d cameras," *IEEE transactions on robotics,* vol. 33, no. 5, pp. 1255-1262, 2017.

[6] C. Yu *et al.*, "DS-SLAM: A semantic visual SLAM towards dynamic environments," in *2018 IEEE/RSJ International Conference on Intelligent Robots and Systems (IROS)*, 2018: IEEE, pp. 1168-1174.

[7] V. Badrinarayanan, A. Kendall, and R. Cipolla, "Segnet: A deep convolutional encoder-decoder architecture for image segmentation," *IEEE transactions on pattern analysis and machine intelligence,* vol. 39, no. 12, pp. 2481-2495, 2017.

[8] B. Bescos, J. M. Fácil, J. Civera, and J. Neira, "DynaSLAM: Tracking, mapping, and inpainting in dynamic scenes," *IEEE Robotics and Automation Letters,* vol. 3, no. 4, pp. 4076-4083, 2018.

[9] K. He, G. Gkioxari, P. Dollár, and R. Girshick, "Mask r-cnn," in *Proceedings of the IEEE international conference on computer vision*, 2017, pp. 2961-2969.

[10] A. Li, J. Wang, M. Xu, and Z. Chen, "DP-SLAM: A visual SLAM with moving probability towards dynamic environments," *Information Sciences,* vol. 556, pp. 128-142, 2021.

[11] W. Wu, L. Guo, H. Gao, Z. You, Y. Liu, and Z. Chen, "YOLO-SLAM: A semantic SLAM system towards dynamic environment with geometric constraint," *Neural Computing and Applications,* vol. 34, no. 8, pp. 6011-6026, 2022.

[12] S. Han and Z. Xi, "Dynamic scene semantics SLAM based on semantic segmentation," *IEEE Access,* vol. 8, pp. 43563-43570, 2020.

[13] Y. Fan, Q. Zhang, Y. Tang, S. Liu, and H. Han, "Blitz-SLAM: A semantic SLAM in dynamic environments," *Pattern Recognition,* vol. 121, p. 108225, 2022.

[14] C. Campos, R. Elvira, J. J. G. Rodríguez, J. M. Montiel, and J. D. Tardós, "Orb-slam3: An accurate open-source library for visual, visual–inertial, and multimap slam," *IEEE Transactions on Robotics,* vol. 37, no. 6, pp. 1874-1890, 2021.

[15] Y. Wu, A. Kirillov , F. Massa, W.-Y. Lo, and R. Girshick. "Detectron2." https://github.com/facebookresearch/detectron2 (accessed 2022).